\documentclass[sigconf,review]{acmart}
\renewcommand\footnotetextcopyrightpermission[1]{}
\makeatletter
\renewcommand\acmConference[3][]{}
\makeatother

\usepackage{graphicx} % Required for inserting images

\title{Fighting AI with AI: Leveraging Foundation Models for  Assuring AI-Enabled Safety-Critical Systems}
\author{Anastasia Mavridou, KBR Inc., NASA Ames\\
Divya Gopinath, KBR Inc., NASA Ames \\
Corina S. P\u{a}s\u{a}reanu, KBR Inc., NASA Ames}
\date{October 2025}

\begin{document}

\begin{abstract}
The integration of AI components, particularly Deep Neural Networks (DNNs), into safety-critical systems such as aerospace and autonomous vehicles presents fundamental challenges for assurance. The opacity of AI systems, combined with the semantic gap between high-level requirements and low-level network representations, creates barriers to traditional verification approaches. These AI-specific challenges are amplified by longstanding issues in Requirements Engineering, including ambiguity in natural language specifications and scalability bottlenecks in formalization. We propose an approach that leverages AI itself to address these challenges through two complementary components. \textbf{REACT} (Requirements Engineering with AI for Consistency and Testing) employs Large Language Models (LLMs) to bridge the gap between informal natural language requirements and formal specifications, enabling early verification and validation. \textbf{SemaLens} (Semantic Analysis of Visual Perception using large Multi-modal models) utilizes Vision Language Models (VLMs) to reason about, test, and monitor DNN-based perception systems using human-understandable concepts. Together, these components provide a comprehensive pipeline from informal requirements to validated implementations.  
\end{abstract}

\maketitle
\pagestyle{plain}

\section{Introduction}

The rise of AI-enabled systems has created a critical challenge. As AI components, such as Deep Neural Networks (DNNs), become integrated into aerospace, autonomous vehicles, and other safety-critical domains, their opacity and complexity create fundamental barriers to assurance. Unlike traditional systems whose behavior can often be tested or formally verified, AI systems exhibit emergent behaviors that resist conventional verification and validation approaches. This opacity is compounded by a semantic gap: requirements are typically expressed in high-level natural language descriptions (e.g., English text), while DNNs process low-level representations (e.g., raw pixels). This mismatch between specification abstraction and implementation creates barriers to standard software engineering practices such as testing, debugging, runtime monitoring, and verification.

These new AI-specific challenges compound the difficulties already inherent to traditional Requirements Engineering (RE) practices. Requirements, typically expressed in natural language by practitioners, are prone to ambiguity, incompleteness and inconsistency~\cite{rozier2016specification}, issues that intensify when specifying requirements for complex, heterogeneous systems integrating AI with conventional components. %The complexity and heterogeneity of such systems demand greater specification effort.
Moreover, requirements for learning-enabled components must extend beyond traditional specifications to capture uncertainty, confidence thresholds, and safety boundaries for emergent behaviors that may arise during operation.

Below, we elaborate on the key challenges that this work seeks to address.

\begin{figure}[t]
    \centering
    \includegraphics[width=1\linewidth]
    {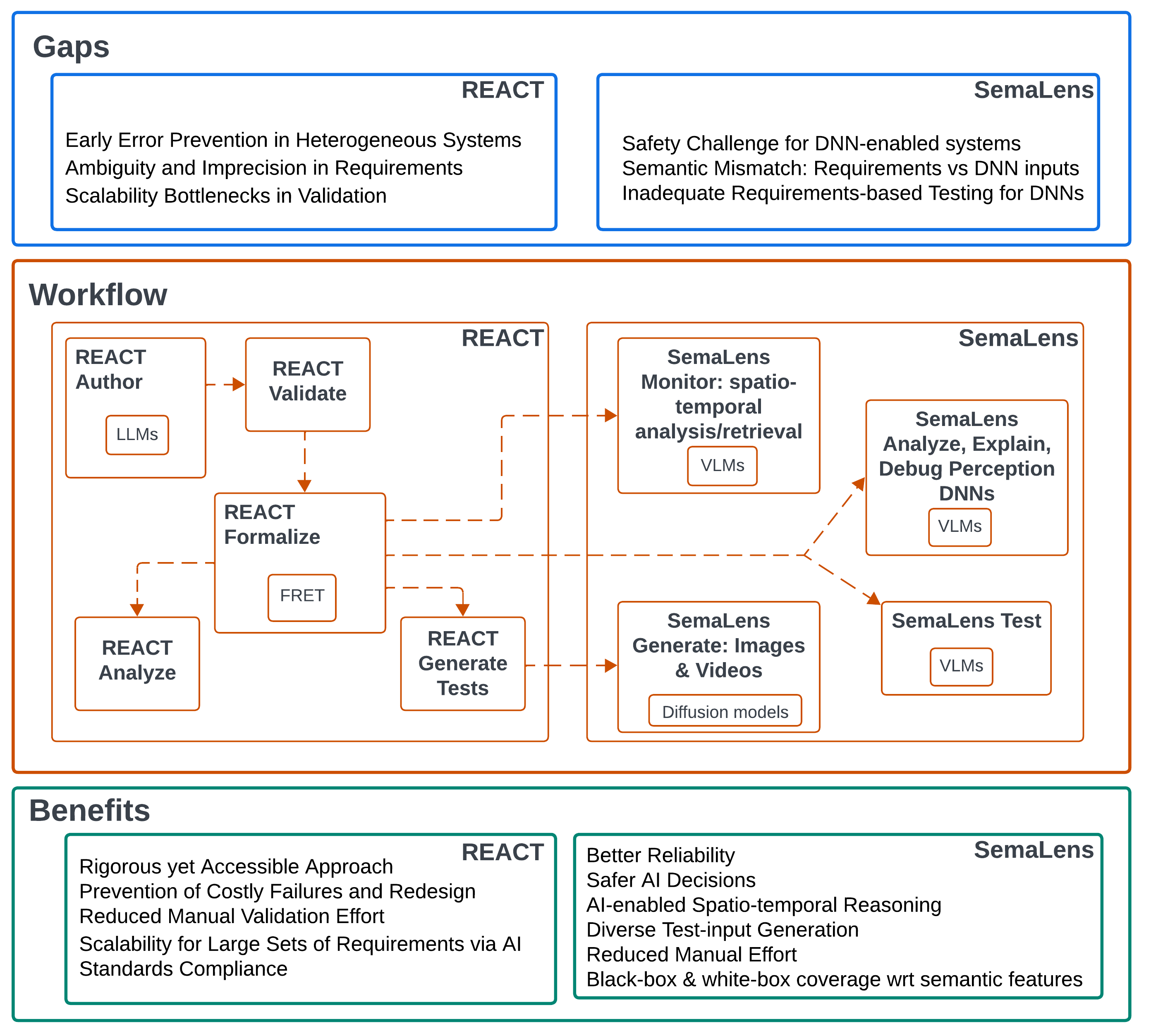}
    \caption{Integrated Framework  with REACT and SemaLens.}
    \label{fig:overview}
\end{figure}

\noindent{\textit{Need for Early Error Detection in Complex, Heterogeneous Systems:}} As systems grow in complexity and heterogeneity, especially with the inclusion of AI, the need for detecting errors and inconsistencies early in the design process becomes critical. If left unaddressed, these issues can propagate into implementation, leading to costly failures and late-stage redesigns. This necessity is especially acute in safety-critical contexts, where requirement errors discovered during operation have resulted in catastrophic failures~\cite{albee2000report}.

\noindent{\textit{Ambiguity and Imprecision in Requirements:}} Effective early detection fundamentally depends on the quality of requirements. Requirements must be clear, unambiguous and precise. Yet, in practice, most requirements are written as natural language statements (e.g., English text), which are inherently ambiguous and prone to misinterpretation~\cite{rozier2016specification}. This creates a major hurdle for designers  and developers, who seek for a single, verifiable source of truth. Furthermore, to accommodate AI, specifications must evolve to include quantitative descriptions of performance boundaries, uncertainty handling, and confidence levels.  

\noindent{\textit{Scalability Bottlenecks in Requirements Engineering:}} The process of translating these informal, potentially ambiguous natural language statements into precise, verifiable specifications is a challenging, time-consuming, and often error-prone task. The translation step demands significant effort from engineers with expertise in formal specification languages, creating a major scalability bottleneck.

\noindent{\textit{Inadequate requirements-based testing:}} Testing and simulation remain crucial for the reliability and safety of critical systems, where failures can have severe consequences. 
Although various techniques have been developed to create test-suites, requirements-based testing for AI systems, particularly the ones that use neural networks, remains largely unexplored.%while adequacy metrics are still missing.

\noindent{\textit{Semantic mismatch between high-level, natural language requirements and low-level network representations:}} Human-specified requirements are written at a high-level, typically in natural language, however, the  inputs and internal processing logic of DNNs have a low-level representation. For instance, a requirement for an autonomous vehicle could be "Always detect pedestrians", however the perception module sees a series of raw pixels and applies an uninterpretable logic to process them.  This disconnect inhibits traceability of DNN behavior to requirements, posing a challenge for verification. 

\noindent{\textit{Safety-assurance bottleneck:}}  DNNs are notoriously hard to analyze, explain and debug due to their opaque and complex nature. It is specifically challenging to measure test coverage of the behavior of perception models with respect to high-level semantic features of interest. For instance, a perception module in an autonomous vehicle would need to be sufficiently tested against different  environment conditions (weather, time of day, obstacles so on) and unexpected vagaries.

In this work, we propose an approach that \emph{fights AI with AI}. In particular, we harness the linguistic and visual reasoning capabilities of modern foundation models 
to ensure the safety and reliability of AI-enabled systems. We introduce two complementary components that strategically deploy large language models (LLMs) and vision-language models (VLMs) to bridge the gaps between informal requirements, formal specifications, and DNN implementations (Figure~\ref{fig:overview}). \textbf{REACT} (Requirements Engineering with AI for Consistency \& Testing) uses LLMs to bridge the gap between informal natural language requirements and formal specifications, enabling automated consistency checking, test case generation aligned with specification semantics, and early verification and validation (V\&V) with full traceability from requirements to test artifacts. \textbf{SemaLens} (Semantic Analysis of Visual Perception using large Multi-modal models) employs VLMs as an analytical "lens" to reason about, test, and monitor DNN-based perception systems in terms of human-understandable concepts, closing the semantic gap between high-level requirements and low-level network representations (i.e., input images and videos). Figure~\ref{fig:overview} shows our overall approach. We explore in detail the workflow in Section 2 and discuss benefits in Section 3.

%The \textbf{SemaLens} (Semantic Analysis of Visual Perception using large Multi-modal models) project comprises of four main research thrusts. \textbf{SemaLens Monitor} leverages vision-language models to perform spatial and temporal reasoning over image sequences (and videos), thereby checking if a given LTL property is satisfied. It can be used offline such as parsing through past accident logs to identify sequences of interest, or can be deployed for runtime monitoring. \textbf{SemaLens Img Generate} uses text-conditional diffusion models to generate semantically diverse images to test the feature-robustness of perception models. \textbf{SemaLens Test} employs the vision-language model (CLIP~\cite{}) to evaluate the coverage of test-suites with respect to high-level features. \textbf{SemaLens Analyze, Explain and Debug} uses CLIP to reason about the logic and behavior of a separate vision model. 

Together, these components offer a comprehensive approach to requirements engineering and verification and validation for AI-enabled safety-critical systems. 
%REACT ensures that requirements are precise, consistent, and verifiable early at design, while SemaLens bridges the semantic gap between natural language specifications and DNN implementations, particularly for perception.
Both leverage AI to achieve scalability while maintaining the rigor necessary for high-assurance systems and compliance with industry standards such as DO-178C. The integration of these components creates an end-to-end pipeline from informal requirements to validated, tested implementations, enabling early error detection, reduced manual effort, and ultimately, safer autonomous systems.
%Figure~\ref{fig:overview} shows our overall approach. We explore in detail the workflow in Section 2 and discuss benefits in Section 3.

\section{Proposed Solution}

\begin{figure*}
    \centering
\includegraphics[width=0.7\linewidth]{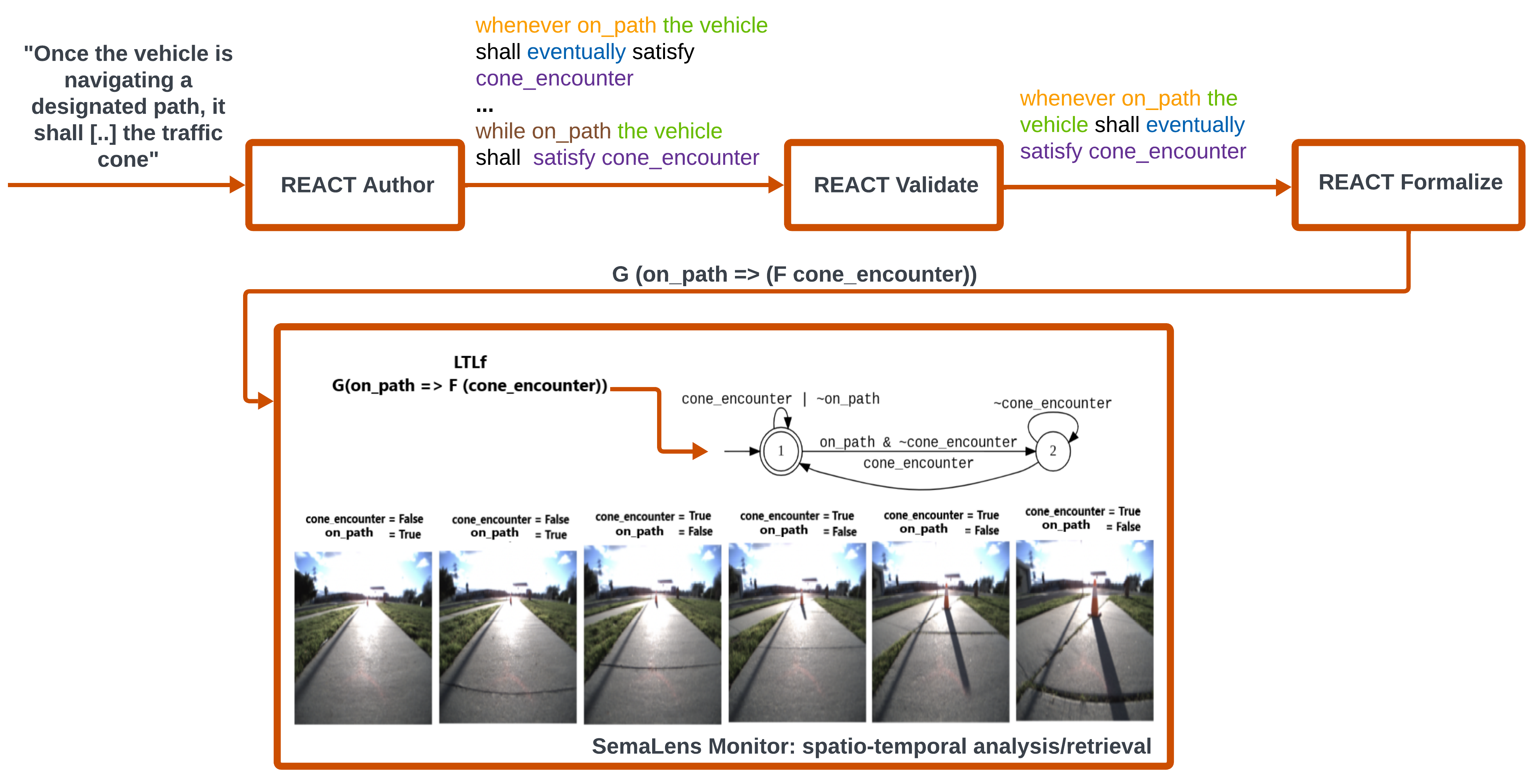}
    \caption{Example  workflow from a natural language requirement to monitoring.}
    \label{fig:semalensmon}
\end{figure*}

Our proposed approach has two components, \textbf{REACT} and \textbf{SemaLens} designed to be complementary and form a rigorous toolchain for the assurance of complex, heterogeneous safety-critical systems. %This integration closes the \emph{assurance loop}. REACT defines \emph{what} the system must safely do and SemaLens provides the capability to test and monitor AI components to confirm \emph{how} they achieve (or violate) the requirements in operation.

%We use the example of Figure~\ref{fig:semalensmon} to demonstrate several steps of our proposed framework. 
Figure~\ref{fig:semalensmon} presents a demonstration of a sample workflow through the framework.
%We use 
Let us consider a requirement that ensures path completion of an autonomous vehicle; for this example we consider an experimental rover developed at NASA\footnote{\url{https://ntrs.nasa.gov/api/citations/20250004071/downloads/RRAV_2025AmesParternshipDays.mp4}}. The initial requirement text written in plain English is the following:

\textbf{[REQ-LIV-002]: }\textit{``Once the rover is navigating a designated path, it shall continue to move and successfully complete the segment by reaching the traffic cone, i.e., the rover must demonstrate non-blocking behavior toward this goal.''}

The requirement states that whenever the rover enters and maintains a state defined as being on the path, it is guaranteed that it will, at some future point, successfully detect and arrive at a target, e.g., a traffic cone. The rover is prohibited from entering an infinite loop, deadlock state, or any other condition that prevents progress towards the cone while navigating on a path.

%{\bf \color{red} I think there is some repetition here: consider removing the paragraphs below or move to intro}
%The REACT component mainly operates at the design and requirements phase, leveraging LLMs to transform ambiguous natural language statements into precise, formal, and verifiable requirements. These formalized requirements directly feed the second framework, to guide the selection of human-understandable concepts and predicates. SemaLens  aims to link these natural-language concepts to the low-level representations the networks operate on (i.e., images or videos). It operates throughout the development lifecycle, starting with requirements-based testing, i.e., computing coverage of  multi-dimensional input data and also generating new data (constrained by natural language requirements) before training, continuing with debuging, explaining, and analyzing trained DNNs wrt {\em semantically meaningful properties}, and further providing run-time monitoring of deployed networks wrt formalized requirements.

Next, we describe in detail the modules of each component.

\subsection{REACT}
The \textbf{REACT} component is an  AI-based Requirements Assistant that aims to help users systematically translate ambiguous natural language requirements  into precise, verifiable formal specifications. By combining the generative capabilities of LLMs with the rigor of formal methods, \textbf{REACT} aims to support early consistency analysis, conflict detection, and automated test case generation directly from requirements. Next, we describe its integrated modules. 

The \textbf{REACT Author} module assists users in authoring precise requirements at scale by combining the linguistic strengths of LLMs with the rigor of formal semantics. It takes as input unrestricted natural language (plain English) text and generates structured natural language (Restricted English (RE))~\cite{prob_spec,10.1007/11663430_6,PSP,fret}, which is a human-readable language with constrained grammar that guarantees unambiguous interpretation. 

For example, in Figure~\ref{fig:semalensmon} (step 1), the natural language requirement \textbf{[REQ-LIV-002]} is fed to REACT Author, which uses LLMs to generate multiple candidate RE requirements translations. Rather than producing a single RE output, the module often produces a list of RE candidates. This is a deliberate choice that stems from the inherent ambiguity of the English requirement. Since such a requirement can often be interpreted in multiple ways, the LLM is tasked with explicitly enumerating all these potential interpretations, allowing users to validate and select the intended meaning.  This approach preserves readability while embedding semantic precision directly into the requirements authoring process. 

The \textbf{REACT Validate} module ensures the correctness of generated requirements by helping users select candidates that match the intended semantics. Since LLMs often produce multiple RE candidates from a single plain English requirement, each with subtle but meaningful semantic differences, the module uses formal validation to automatically distinguish these variations. Rather than exposing users to complex formal logic, semantic differences are presented in engineer-friendly formats (e.g., execution traces or concrete scenarios). By simply accepting or rejecting a semantic difference, users can efficiently prune incorrect candidates. For example, as shown in Figure~\ref{fig:semalensmon}, in the end, the pruning process yields a single RE requirement that corresponds to the user's intended semantics. The validation process is highly targeted, focusing only on key semantic distinctions to minimize the manual effort required. This human-in-the-loop validation is critical. It ensures that requirements are vetted against the user's actual intent rather than relying on AI interpretation, which may misunderstand domain-specific nuances.

The \textbf{REACT Formalize} module translates validated RE requirements into formal specifications through seamless integration with requirement formalization tools such as FRET~\cite{giannakopoulou2020formal}. The module generates formal representations in formal logics such as Linear Temporal Logic for finite traces (LTLf)~\cite{10.5555/2540128.2540252}, as shown in Figure \ref{fig:semalensmon}. 
To handle the complexity and uncertainty inherent in autonomous systems, this module supports translation to formal logics that accommodate requirement types specific to AI components, including Vision-Language Model (VLM)-based perception systems. This capability is essential for formalizing requirements that capture uncertainty, specify confidence criteria and bound emergence. 

The \textbf{REACT Analyze} module supports robust early verification and validation (V\&V) by performing automated formal analysis across the requirement set. This capability systematically detects inconsistencies and conflicts at design time, before implementation begins, reducing the need for costly downstream rework and preventing defects from propagating into code.

Finally, the \textbf{REACT Generate Test Cases} module leverages the formalized requirements to automatically produce candidate test cases with coverage guarantees, which aims at addressing the requirements-based testing and coverage objectives of DO-178\footnote{Note that DO-178 was not designed for the assurance challenges presented by Learning-Enabled Components like DNNs. However, the reality in aerospace (and other safety-critical domains) is that autonomous platforms often consist of mixed AI and traditional components. Therefore, DO-178 remains the mandatory baseline requirement for all non-AI, safety-critical software embedded in the system. 

Addressing the complexity of the integrated AI components requires leveraging emerging standards and guidelines specifically designed for assurance under uncertainty. For aerospace, a key example is the guidance emerging from the SAE G-34 working group, which is focused on defining certification and assurance methodologies for AI in Aviation to complement DO-178.}. 

This module accelerates testing workflows and ensures comprehensive, requirement-driven validation. It guarantees complete traceability by creating an explicit link between every formal requirement and its corresponding test cases. These test cases, can then be given as input to SemaLens, to generate videos that check the semantic robustness of perceptions models.

\subsection{SemaLens}
The \textbf{SemaLens} component of our framework aims to leverage emerging multi-modal foundation models such as vision-language models to analyze, test, debug, explain, and monitor DNNs used in perception for autonomous systems in terms of human understandable concepts. %; the approach also aims to leverage VLMs to build monitors for off-line and on-line spatio-temporal reasoning over sequences of images and videos and to explore text-conditional diffusion models to generate semantically-diverse test images and videos for testing perception modules.  
Vision-Language Models (VLMs) \cite{VLMsurvey}, such as CLIP~\cite{pmlr-v139-radford21a}, are powerful models trained on massive amounts of images and textual data and can thus serve as a rich repository of human-understandable concepts for diverse images.
%be queried on vision and language modalities.
%For instance CLIP~\cite{pmlr-v139-radford21a} is a model trained on a large corpus of images accompanied by captions describing the images. Our insight is to leverage these models to assign semantics to images in terms of natural language descriptions with human-understandable concepts.

%{\bf corina TODO: explain that VLMs were trained on many images and test captions so we can leverage them to assign semantics to images in terms of natural language descriptions/concepts} We describe the various modules of SemaLens below.

\noindent{\textbf{SemaLens Monitor:}}
This module enables spatial and temporal reasoning over sequences of images (and videos). It uses VLMs to extract concepts and spatial relationships from individual images and uses temporal logic to capture temporal relationships between images in a sequence, thereby building a monitor for automatic analysis of videos and image sequences. 

The monitor could be used \textit{offline} to find “unusual” sequences of (unlabeled) images/videos; e.g., parse through collection of past accident videos and select sequences corresponding to risk scenarios. It can also be deployed \textit{online} to check image sequences/videos at run-time for conformance with requirements and flag deviations, particularly useful in safety-critical settings. Although VLMs are still struggling to extract complex spatial relationships from images, previous work, e.g.,~\cite{11135768}, has shown promising results.

\textit{SemaLens Monitor} finds a natural integration with \textit{REACT}. % \textit{REACT Formalize} can synthesize formal temporal properties from natural language requirements, which can in turn be fed into \textit{SemaLens Monitor} to automatically synthesize monitors.
Consider the example in Figure~\ref{fig:semalensmon}. \textit{REACT Formalize} synthesizes the property in LTLf from the given requirement in English. The formula is parsed and converted to a Deterministic Finite Automaton (DFA) (note that symbols: `|', `\&', `$\sim$' denote: logical {\em or, and, negation}, respectively). To evaluate the automaton on a sequence of images, the predicates $on\_path$ and $cone\_encounter$ need to be evaluated on each image. This is done by feeding the image through the CLIP model (ViT-B/16) and computing the similarity~\cite{radford2021clip} between the respective image embedding and the embeddings of textual captions corresponding to the predicates. In the example, a predicate is evaluated to True if the respective similarity is greater than a threshold (0.4 in our case). Figure~\ref{fig:semalensmon} shows a sequence of images with respective predicate evaluations which satisfy the property; the monitor returns True from the third image onward.

\noindent{\textbf{SemaLens Img Generate:}}
This module aims to use text-conditional diffusion models to generate semantically diverse test images (and videos) that conform with requirements written in natural language. It generates test images/videos constrained by input preconditions.
%, which further contain a diversity in features (that could not be present in simulators). 
While previous work has explored text-conditional diffusion models for requirements-based test image generation, see e.g.,~\cite{mozumder2025rbt4dnnrequirementsbasedtestingneural}, we plan to build on that work with further adding semantic perturbations conditioned on prompts.  The resulting test-suites can serve to check semantic robustness of perception models.

The module can be integrated with \textit{REACT Generate Tests} to take test sequences that satisfy temporal specifications and use them to generate videos; the test sequences ensure requirement coverage, while the diffusion model helps cover semantic features.
%(environmental vagaries).

\noindent{\textbf{SemaLens Test:}}
This module aims to use VLMs to %{\em extract} high-level semantic features from images and 
define novel {\em coverage metrics} over sets of (unlabeled) images in terms of semantic features (that can for instance appear in the operational design domain -- ODD -- of the autonomous system). An image is said to cover a feature if the similarity score between the image embedding (as computed by the VLM) and the textual embedding of the feature is higher than a user-specified threshold. Statistical measures can be used to quantify how {\em well} a feature is covered by the set of images.

This capability enables both {\em black-box} and {\em white box} testing; in black-box mode we analyze an (unlabeled) data set to compute coverage of relevant features and identify gaps; in white-box mode we first {\em map} the embedding space of a perception component to the embedding space of a VLM and compute coverage through the lens of the VLM.

\noindent{\textbf{SemaLens AED (Analyze, Explain, and Debug):}}
This module uses a VLM as a lens to reason about the logic and behavior of a separate vision model. The crux of the approach is to build a {\em map} that aligns the embeddings of a vision model with the internal representations of the CLIP model \cite{11030022,DBLP:conf/saiv/MangalNGHRJP14,moayeri2023texttoconceptandbackcrossmodel}. The embedding space of CLIP thereby acts as a proxy to analyze, explain, and debug the DNN model's behavior in terms of user-defined concepts (without requiring manual annotations).

For instance, if a model classifies an image as a \textit{truck}, it can be %checked 
examined if it is doing so for the right reasons, by checking if relevant concepts such as \textit{metallic} and \textit{rectangular} are also detected by the model.
%more strongly than irrelevant concepts such as \textit{eyes} and \textit{ears}. The work in  define a formal language that enables specifying such properties as \textit{strength predicates}, evaluated as mathematical constraints in CLIP's embedding space. 
Such concepts can act as semantically-meaningful explanations for the  behavior of  the perception module. Furthermore, if an image is misclassified by the model, the semantic mapping can be used to localize if the bug lies in the vision \textit{encoder} leading to wrong concepts being extracted or in the logic of the \textit{head}. 

The behavior of the model with respect to different concepts can be analyzed statistically, to obtain {\em semantic heatmaps}, that %succinctly summarize the strength of different concepts.  These heatmaps 
enable the identification of non-robust and brittle features.
%, whose predicates are frequently violated in errors. 
The heatmaps could also be deployed at runtime to flag adversaries and unsafe inputs, when their semantic profile deviates from expected patterns. 
%Refer \cite{11030022,DBLP:conf/saiv/MangalNGHRJP14} for preliminary work on these ideas.

%This idea is leveraged to develop debugging techniques to localize and explain layers/neurons responsible for defects/misclassifications in terms of human-understandable concepts. The localized concepts could in turn can be used to select images and videos for targeted repair/retraining. The capability could also be employed to flag adversaries and unsafe inputs at runtime when their semantic profile deviates largely from expected concept-based patterns. Refer \cite{11030022, DBLP:conf/saiv/MangalNGHRJP14} for preliminary work on these ideas.

Both \textit{SemaLens AED} and \textit{SemaLens Test} can be integrated with \textit{REACT} to obtain the vocabulary of high-level human-understandable concepts of interest.%, built during requirements elicitation or obtained from operational design documents.

\section{Benefits}
We summarize below the benefits enabled by our framework.% which elucidate that it can play a key role in contributing towards safety assurance of autonomous systems.

\subsection{REACT}
\textbf{Rigorous yet Accessible Approach:} REACT combines the prevision of formal analysis with the usability of pure English, offering a practical pathway for users to improve requirement quality without demanding expertise in formal specification. Provides execution traces, interpretable differences, and rationale for detected issues -- not opaque outputs.\\
\textbf{Early Verification and Validation to Prevent Costly Failures and Redesigns:} Designed to enable early Verification and Validation (V\&V) directly from English requirements at the earliest stages of design. It catches ambiguities and inconsistencies before they propagate downstream, reducing costly late-stage fixes.\\
\textbf{Reduced manual validation effort:} Our validation approach focuses on key semantic distinctions, which reduces manual effort.\\
\textbf{Scalability to Complex Projects through AI:} Leverages LLMs to process and reason over large sets of requirements, making it suitable for time-sensitive, high-assurance systems.\\
\textbf{Standards Compliance:} Facilitates alignment with industry standards, such as DO-178C.

\subsection{SemaLens}

\noindent{\textbf{Better reliability:}} Runtime Analysis detects and mitigates AI errors by checking conformance to requirements in real-time thereby improving system reliability.

\noindent{\textbf{Safer AI decisions:}}
Explains what AI “sees” and why it makes certain decisions contributing to safer autonomy.

\noindent{\textbf{Reduced manual effort in debugging:}} Interprets and debugs without requiring costly, time-consuming human annotations. 

\noindent{\textbf{AI-enabled spatial-temporal reasoning}}: Enables complex reasoning  over multiple modalities (image/text).

\noindent{\textbf{Generation of diverse test inputs}}: Enables testing in unusual scenarios (beyond simulation) and for data-sparse environments to ensure robustness and safety of autonomous systems with DNN perception. 
%\end{itemize}

\noindent{\textbf{Black-box and white-box coverage wrt high-level features}}: Proposes novel coverage metrics for image domain without expensive human annotations.

\section{Conclusion}
This research idea paper describes emerging ideas on providing safety-assurance for autonomous systems by exploring novel methods to leverage AI in various life cycle stages. We proposed a workflow that incorporates two synergistic components, \textbf{REACT} and \textbf{SemaLens}, which offer a comprehensive approach harnessing multi-modal capabilities of foundation models to enable requirements engineering, verification, validation and monitoring for AI-enabled safety-critical systems. By fighting AI with AI, our approach strives to address challenges that currently prevent certification of learning-enabled components in safety-critical applications.
\newpage
\bibliographystyle{plain}
\bibliography{main}

\end{document}